%% file: new.tex
\documentclass{article} 
\usepackage{iclr2027_conference,times}

\input{math_commands.tex}

\usepackage{graphicx}
\usepackage{hyperref}
\usepackage{url}
\usepackage{booktabs}
\usepackage{multirow}
\usepackage{wrapfig}

\usepackage{amsmath}
\usepackage{amssymb}
\usepackage{mathtools}
\usepackage{amsthm}

\usepackage{booktabs} 
\usepackage{bm}
\usepackage{stfloats}
\usepackage{enumitem}

\title{TabGift: A Context-Efficient Foundation Model for Tabular Data Generation}

\author{Yu-Sheng Li\footnotemark[1] \quad
Jun-Peng Jiang\footnotemark[1] \quad
Han-Jia Ye\footnotemark[2] \\
School of Artificial Intelligence, Nanjing University, China \\
National Key Laboratory for Novel Software Technology, Nanjing University, China \\
\texttt{\{liys,jiangjp,yehj\}@lamda.nju.edu.cn}}

\iclrfinalcopy 
\begin{document}

\maketitle
\begingroup
\renewcommand{\thefootnote}{}
\footnotetext[1]{\textsuperscript{*}Equal contribution, \textsuperscript{$\dagger$}Corresponding author.}
\endgroup

\begin{abstract}
Tabular data generation aims to synthesize realistic samples from limited observations while faithfully capturing the underlying data distribution.
Tabular foundation models (TFMs), pretrained on diverse tabular tasks and adapted through in-context learning, have demonstrated strong predictive capabilities, suggesting the potential of extending this paradigm to tabular data generation.
However, the strong predictive performance of TFMs often relies on sufficiently large context sets, whereas tabular generation is typically needed precisely when target data are scarce. Extending the foundation-model paradigm to generation therefore requires reliable distribution inference from only a limited context.
To this end, we propose {\sc TabGift}, a \textit{context-efficient} tabular foundation model enabling both efficient distribution modeling and effective generation from only a few context samples. {\sc TabGift} maps heterogeneous tables into a shared latent space with a pretrained tabular encoder, where an in-context diffusion model learns to denoise latent samples conditioned on a small set of clean examples. This design transfers generative priors across datasets and enables distribution inference from limited context, while a lightweight dataset-specific MLP decoder maps generated representations back to the original data space.
Experiments on 7 classification and 11 regression datasets show that {\sc TabGift} consistently outperforms representative baselines in synthetic data quality, with particularly strong gains under limited-context settings, while achieving substantially faster adaptation and generation.

\end{abstract}

\input{section_intro}

\input{section_relatedwork}

\input{section_method}

\input{section_pretraining}

\input{section_experience}

\section{Limitations} 
SwiftPFN is pretrained for prediction, so its embeddings are not necessarily optimal for representing a sample in generation, and a dataset-specific MLP decoder is still required. We have explored alternatives such as a universal decoder and jointly training the encoder and decoder as a general tabular VAE, but both yielded inferior results. More systematic efforts are needed to design an encoder–decoder pair that is both transferable and generation-oriented.
 
Pretraining on synthetic data alone leads to suboptimal generation quality, and adding real data improves performance, suggesting that the synthetic prior is insufficiently diverse or realistic. Future work could improve the quality of synthetic pretraining data, scale up real-data pretraining, and increase per-dataset size and feature dimensionality.

\section{Conclusion}
{\sc TabGift} is an efficient foundation model for tabular data generation, designed to address the key limitations of existing generative methods in small-data regimes, including overfitting, poor privacy protection, and high computational cost. 
Its core idea is to shift the burden of distribution learning from the target dataset to foundation-model pretraining: by pretraining on large-scale heterogeneous tabular data, {\sc TabGift} acquires transferable structural and distributional priors, enabling high-quality synthetic data generation from only a few context samples.
{\sc TabGift} has three components: a frozen SwiftPFN encoder mapping heterogeneous samples to a 192-d latent space; an 18-layer Transformer denoiser with a sample-wise attention mask that enables context-guided denoising while keeping query denoising independent; and a lightweight dataset-specific MLP decoder mapping latents back to the data space.
Training proceeds in two stages: large-scale pretraining on synthetic data (50,000 batches), followed by fine-tuning on 72 real-world datasets (10,000 batches). Experiments on 7 classification and 11 regression tasks show that {\sc TabGift} consistently achieves the best or near-best generation quality in the limited-context regime (size 16–64), while being substantially more efficient than most baselines. 


\newpage
\section*{AI Use Statement}
In this work, we used generative AI tools for writing assistance, including polishing and rephrasing text, and for coding assistance, including debugging and implementing standard routines. We have not used generative AI tools for research ideation, problem formulation, or methodological design. These aspects, including the core idea, method, and experimental design, were conceived and carried out by the authors.

Additionally, we used generative AI tools for literature search and other auxiliary tasks such as summarizing related work and checking terminology. We have reviewed all AI-assisted work. LLM-generated code was verified and tested for correctness by the authors, and AI-assisted text was carefully checked and revised. We take responsibility for the final content of this work, including text, claims, or artifacts produced with the aid of generative AI.

\section*{Ethics Statement}
This work does not involve human subjects, personal data collection, or any interaction with human participants. All datasets used in our experiments are publicly available and were obtained from standard open benchmarks. We did not collect, release, or redistribute any new datasets.

\section*{Reproducibility Statement}
We have made efforts to ensure the reproducibility of our work. Detailed descriptions of the model architecture, training procedure, and experimental setup are provided in the main paper and the appendix. All datasets used in our experiments are publicly available, and the corresponding preprocessing and sampling procedures are described in the appendix. We intend to release the pretrained model together with the accompanying training and inference code upon publication, which will allow the results reported in this work to be reproduced. The authors are committed to releasing these resources in a timely manner.

\bibliography{iclr2027_conference}
\bibliographystyle{iclr2027_conference}

\newpage
\appendix
\input{appendix}

\end{document}

%% file: math_commands.tex
\usepackage{amsmath,amsfonts,bm}

\def\eqref#1{equation~\ref{#1}}

\def\1{\bm{1}}

\def\vx{{\bm{x}}}
\def\vy{{\bm{y}}}

\def\mX{{\bm{X}}}

\def\mZ{{\bm{Z}}}

\DeclareMathAlphabet{\mathsfit}{\encodingdefault}{\sfdefault}{m}{sl}
\SetMathAlphabet{\mathsfit}{bold}{\encodingdefault}{\sfdefault}{bx}{n}

\newcommand{\E}{\mathbb{E}}



%% file: section_intro.tex
\section{Introduction}


Tabular data are ubiquitous in real-world applications such as finance~\citep{zheng2020pcal, clements2020sequential, huang2023aligning}, healthcare~\citep{gong2020evaluation, yang2021flop}, and scientific discovery~\citep{du2021tabularnet, wu2021fedgnn, zhang2023lightfr}. However, high-quality tabular data are often limited by privacy constraints, restricted accessibility, and costly collection or annotation. Tabular data generation~\citep{mehrabi2021survey, dastin2022amazon} therefore provides an important means to alleviate data scarcity by synthesizing realistic samples that preserve the underlying data distribution while reducing reliance on direct access to sensitive or limited real-world data.

Existing tabular generation methods are mainly non-parametric~\citep{chawla2002smote, he2008adasyn, zhang2018mixup} or parametric~\citep{xu2019modeling, shi2024tabdiff, zhang2024mixed}. The former depend on observed samples for resampling or interpolation, while the latter require dataset-specific training and tuning. These limitations motivate tabular foundation models, which aim to amortize learning across datasets and adapt to new tables with minimal target-specific optimization.

Tabular foundation models~\citep{hollmann2025accurate, jaeger2026tabpfn35technicalreport, grinsztajn2026tabpfn3technicalreport, grinsztajn2025tabpfn} provide a promising solution to these limitations. By pretraining on diverse tabular tasks and adapting through in-context learning, they can achieve strong predictive performance on unseen datasets without dataset-specific retraining or hyperparameter tuning. Their ability to rapidly transfer across datasets naturally suggests a promising direction for tabular generation, leading to a direct question: \textit{can tabular foundation models also be extended to tabular data generation?}

However, extending TFMs from prediction to generation introduces a distinct challenge in context efficiency. Unlike prediction, which only requires task-specific outputs conditioned on observed features, generation must infer the joint distribution of all variables from the provided context, making reliable generation more demanding under limited observations. Yet relying on large context sets is fundamentally at odds with the data-scarce settings~\citep{wang2025rdbpfn, park2026tabular, lin2026retabsyn, shen2026labtab} where synthetic data are most valuable. Meanwhile, accurately capturing dependencies within the context often requires fine-grained interactions among features and samples, whose computational cost grows rapidly with context length and table dimensionality~\citep{dpganit2025}. Therefore, an effective generative TFM must not only infer target distributions from a small context, but also model that context efficiently.

To address these challenges, we propose {\sc TabGift}, a \textit{context-efficient foundation model} for tabular data generation. {\sc TabGift} builds a shared latent space to unify heterogeneous tables, performs in-context generation in this space to infer target distributions from limited observations, and uses lightweight dataset-specific decoding for efficient adaptation. Together, these designs enable high-quality generation from short contexts with low computational overhead.

\begin{figure}[t]
  \centering
  \includegraphics[width=\textwidth]{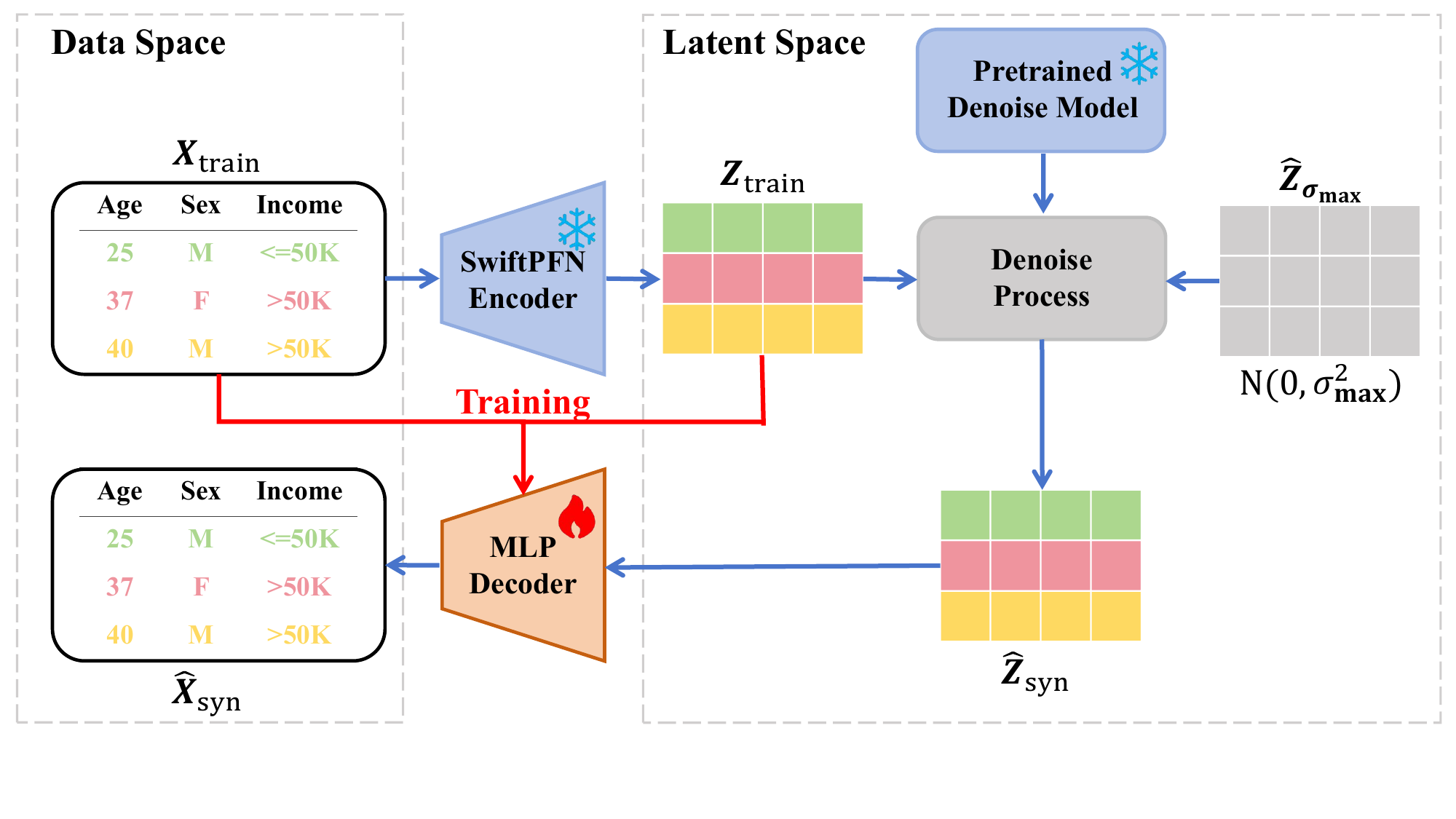}
  \vspace{-2em}
  \caption{{\sc TabGift} consists of a data space and a latent space:
  (1) Encoding: Given the real training table $X_\text{train}$, the SwiftPFN Encoder maps heterogeneous samples into a unified 192-dimensional latent space, producing the latent embeddings $Z_\text{train}$.
  (2) Latent-space generation: The real-data latent latent embedding $Z_\text{train}$ is used as context. Starting from Gaussian noise, a pretrained frozen denoising model generates new latent embeddings $\hat{Z}_\text{syn}$ through an iterative denoising process.
  (3) Decoding: A lightweight MLP Decoder is trained per target dataset to reconstruct the generated latent embeddings $\hat{Z}_\text{syn}$ back into the data space, yielding synthetic data $\hat{X}_\text{syn}$.
  Throughout the pipeline, only the MLP Decoder requires dataset-specific training, while the SwiftPFN Encoder and the denoising model are pretrained and kept frozen during inference.}
  \vspace{-1em}
  \label{fig:fig1}
\end{figure}

The generation pipeline is illustrated in Fig~\ref{fig:fig1}. Specifically, {\sc TabGift} first employs a pretrained tabular encoder to map rows from heterogeneous tables into a shared latent space, enabling a unified generative model across datasets with different schemas. In this space, we develop an in-context diffusion model that conditions noisy latent samples on a small set of clean context examples and performs denoising through efficient row-level interactions. This formulation allows the diffusion model to reuse generative knowledge across datasets while adapting the generated distribution to the provided context. Finally, a lightweight dataset-specific MLP decoder maps the generated latent representations back to the data space, requiring only minimal target-specific adaptation. Experiments on 7 classification and 11 regression datasets show that {\sc TabGift} consistently outperforms representative baselines in synthetic data quality, with particularly strong gains under limited-context settings and substantially improved generation efficiency. Our contributions are threefold:
\begin{itemize}[noitemsep,topsep=0pt,leftmargin=*]
\item We highlight the importance of context efficiency in foundation models for tabular generation, showing that long-context dependence fundamentally conflicts with the data-scarce settings where synthetic data generation is most valuable.
\item We propose {\sc TabGift}, a context-efficient tabular foundation model that combines transferable distributional priors, efficient row-level modeling, and lightweight dataset-specific decoding for fast generation from limited context.
\item We conduct extensive experiments across 18 classification and regression datasets, demonstrating that {\sc TabGift} consistently improves synthetic data quality under limited-context settings while achieving substantially higher generation efficiency.
\end{itemize}

%% file: section_relatedwork.tex
\section{Related Work}
\subsection{Tabular Generation Models}

Early work on tabular data generation primarily focused on improving downstream predictors rather than learning a high-fidelity data generator. Representative methods such as SMOTE~\citep{chawla2002smote} and ADASYN~\citep{he2008adasyn} address class imbalance by synthesizing minority samples, but they do not aim to match the full joint distribution of the original table.

A subsequent line of work instead fits expressive generative models to observed samples. These methods can be broadly grouped into two categories. The first adapts image-inspired generators to tables, including GAN-based models such as ADSGAN~\citep{yoon2020anonymization} and CTGAN~\citep{xu2019modeling}, VAE-based approaches such as TVAE~\citep{xu2019modeling}, and diffusion-based generators such as TabDDPM~\citep{kotelnikov2023tabddpm}, TabDiff~\citep{shi2024tabdiff}, and TabSyn~\citep{zhang2024mixed}. The second develops table-specific families: Bayesian networks~\citep{Ankan2024} factorize the joint distribution along a directed acyclic graph, normalizing flows~\citep{durkan2019neural} provide tractable likelihoods and efficient sampling, and tree-based alternatives such as Adversarial Random Forests~\citep{watson2023adversarial} offer non-parametric density estimation and capture feature interactions. While these approaches can model complex distributions, they are typically trained per dataset and often degrade when training data are scarce.

\subsection{Tabular Foundation Models}
Tabular foundation models have evolved rapidly from the Prior-Fitted Network (PFN) paradigm introduced by TabPFN~\citep{hollmann2023tabpfn, grinsztajn2025tabpfn, hollmann2025accurate, jaeger2026tabpfn35technicalreport, grinsztajn2026tabpfn3technicalreport}, which pretrains on synthetic tasks and performs prediction on unseen datasets through in-context learning. 
Building on this framework, TabICL~\citep{qu2025tabicl} further improves scalability and representation learning by introducing column-aware modeling and a multi-stage pretraining strategy, making foundation models applicable to larger and more heterogeneous tables. 
More recently, the field has expanded considerably, with models such as TabFM~\citep{kong2026tabfm}, EXAONE~\citep{eo2026exaone}, Mitra-2~\citep{tao2026mitrav2}, LimiX~\citep{limix2m2026}, Xiaomi-TabLDM~\citep{wang2026xiaomi}, and newer versions of TabPFN exploring different directions in model scale, task coverage, architectural design, and pretraining strategies. Together, these developments have substantially advanced the capability and generality of tabular foundation models.

Compared with predictive modeling, tabular foundation models for data generation remain relatively underexplored. 
TabPFGen~\citep{ma2023tabpfgen} and TabEBM~\citep{margeloiu2024tabebm} combine TabPFN with energy-based modeling~\citep{grathwohlyour} to achieve foundation model generation ability.
TabFORGE~\citep{jiang2026tabforge} leverages pretrained causality-aware tabular representations for generative modeling, while DiffICL~\citep{han2026difficl} formulates tabular synthesis as an in-context learning problem to infer target distributions from limited examples. Despite these advances, existing methods still rely on relatively long contexts or costly interaction modeling, leaving context-efficient and computationally efficient generation insufficiently explored.

%% file: section_method.tex
\section{Method}
\subsection{Overview}
In this section, we introduce the architecture of {\sc TabGift} and how to use {\sc TabGift} to generate new samples. 
{\sc TabGift} can be viewed as a latent diffusion model (LDM)~\citep{lin2025ctsyn,zhu2025dptldm,correlationloss2025} consisting of three components: (1) an encoder obtained from the general tabular prediction model SwiftPFN~\citep{liu2026swiftpfn}, which leverages its pretrained generalization ability on tables to map heterogeneous tabular data into a unified latent space; (2) a score-based diffusion model~\citep{zhang2024mixed} trained in the latent space via in-context learning, where latent embeddings of real samples serve as context to help the denoising process for noisy embeddings, enabling sampling from random noise; and (3) a dataset-specific MLP decoder that maps the generated latent embeddings from the latent space back to the concrete data space.

\subsection{Setup}
We consider a tabular dataset $X{=}\{\vx_i\}_{i=1}^{N}$, where $N$ is the number of samples, $\vx_i {\in} \mathcal{X}_1 {\times} {\cdots} {\times} \mathcal{X}_F$ is the feature vector with $F$ features. Features may be numerical or categorical.

Given a training set $X_{\text{train}}$, tabular data generation aims to learn a model \(p_\theta\) that produces a synthetic dataset $\hat{X}_{\text{syn}}$ approximating the underlying distribution of $X_{\text{train}}$, while preserving marginal distributions and inter-feature dependencies and avoiding copying of training samples.

\subsection{SwiftPFN Encoder}
We use SwiftPFN as a general-purpose encoder to handle heterogeneous datasets, which is a foundation tabular prediction model. Using only sample-attention, SwiftPFN achieves performance comparable to Limix~\citep{limix2m2026} and TabPFNv2~\citep{hollmann2025accurate} (which use sample-attention and feature-attention). In specific, SwiftPFN first maps both the training samples ($\displaystyle \mX_\text{train}$ and $\displaystyle \vy_\text{train}$) and the test samples ($\displaystyle \mX_\text{test}$ only) to be predicted into the same dimension ($d = 192$). It then uses a 12-layer Transformer architecture with a causal mask (training samples can attend to each other, while test samples can only attend to training samples) to enable interactions between the training and test samples. This allows the embedding of a test sample to learn label information, and finally a prediction head outputs the prediction result. Formally, the prediction process of SwiftPFN can be written as the following equation:
\begin{align*}
\vy_\text{pred}
&= \text{SwiftPFN}(\mX_\text{test};\mX_\text{train},\vy_\text{train}) \\
&= \text{PredHead}\Bigl(\text{Transformer}\bigl([\text{Embed}(\mX_\text{train},\vy_\text{train}], \text{Embed}(\mX_\text{test}))\bigr)[n:]\Bigr),
\end{align*}
where $n$ denotes the length of the training samples, and $[n:]$ denotes taking the embedding corresponding to the test samples.
Here, we can take the sample representation after the multiple Transformer layers and before the prediction layer as the latent embedding we desire. 
Since in generation tasks there is only $\displaystyle \mX_\text{train}$, and we do not distinguish which column is the prediction target, to generate sample embeddings through SwiftPFN, we additionally construct an all-zero label $\displaystyle \vy_\text{train} = \mathbf{0}_n$ and an all-zero test sample $\displaystyle \mX_\text{test} = \mathbf{0}_{1{\times} F}$ to ensure the normal operation of the model. 
\begin{align*}
\mZ_\text{train}
&= \text{SwiftPFNEncoder}(\mathbf{0}_{1\times F};\mX_\text{train},\mathbf{0}_n) \\
&= \text{Transformer}\bigl([\text{Embed}(\mX_\text{train},\mathbf{0}_n), \text{Embed}(\mathbf{0}_{1{\times} F})]\bigr)[:n].
\end{align*}

\subsection{In-Context Diffusion Model}
Diffusion models aim to train a denoiser that recovers the original data from noisy data and the noise level, their theory is already well established. 
{\sc TabGift} trains a context-conditioned denoising network: by taking the latent embeddings of real samples as context, it learns to estimate the denoising direction for noisy latent embeddings, thereby avoiding dataset-specific training.

\subsubsection{Forward Training Process}
We adopt a score-based diffusion process as the main generative framework of {\sc TabGift}, which has been widely validated in tabular generation tasks. Specifically, given the latent embedding $\mZ_\text{train}$ obtained by the SwiftPFNEncoder, we divide it into two parts: $\mZ_\text{clean}$ and $\mZ_\text{noisy}$. $\mZ_\text{clean}$ remains clean and serves as the in-context condition, while $\mZ_\text{noisy}$ is treated as the target to be perturbed with noise of different magnitudes. The noising process is as follows:
$$
\mZ_\sigma = \mZ_\text{noisy} + \sigma\epsilon,\quad\epsilon\sim \mathcal{N} (0, 1),
$$
where $\sigma$ denotes the noise magnitude. It is typically sampled from a log-normal distribution $\ln{\sigma}{\sim} \mathcal{N} (p_\text{mean}, p_\text{std}^2)$. The denoiser takes $\mZ_\sigma$ and the clean context $\mZ_\text{clean}$ as input, and is trained to predict $\mZ_\text{noisy}$. The training objective is
$$
\ell_\text{diff}=\E_{\mZ_\text{noisy},\sigma,\epsilon}\Big[\frac{\sigma^2 + \sigma^2_{\text{data}}}{\sigma^2\sigma^2_{\text{data}}}\big(\text{Denoiser}(\mZ_\sigma, \sigma;\mZ_\text{clean})-\mZ_\text{noisy}\big)^2\Big].
$$
The EDM-parametrized~\citep{karras2022edm} denoiser estimates the clean latent:
$$
\text{Denoiser}(\mZ_\sigma, \sigma;\mZ_\text{clean})=c_\text{skip}(\sigma)\mZ_\sigma + c_\text{out}(\sigma)D_\theta\big(c_\text{in}(\sigma)\mZ_\sigma, c_\text{noise}(\sigma);\mZ_\text{clean}\big).
$$
Network $D_\theta$ is the in-context learning model that truly requires training and $c_\text{skip}(\sigma)$, $c_\text{out}(\sigma)$, $c_\text{in}(\sigma)$, $c_\text{noise}(\sigma)$ are standard EDM preconditioning coefficients.

\subsubsection{Sampling Process}
Sampling is performed in the latent space with $\mZ_\text{train}$ as context. We initialize each entry of $\hat{\mZ}_{\sigma_{\text{max}}}$ from $\mathcal{N} (0, \sigma_{\text{max}}^2)$ and denoise from $\sigma_{\text{max}}$ to $\sigma_{\text{min}}$.
Under the EDM parametrisation, we update:
$$
\hat{\mZ}_{\sigma_{i-1}} = \hat{\mZ}_{\sigma_{i}} + (\sigma_{i-1} - \sigma_{i})\frac{\hat{\mZ}_{\sigma_{i}} - \text{Denoiser}(\hat{\mZ}_\sigma, \sigma_i;\mZ_\text{train})}{\sigma_{i}}.
$$
Among them, $\sigma_{i}$ is a set of preset noise amplitudes from $\sigma_{\text{max}}$ to $\sigma_{\text{min}}$. The generated latent embedding $\hat{\mZ}$ is obtained after multiple steps of denoising.

\subsubsection{In-Context Learning Model}
\begin{wrapfigure}{r}{0.4\textwidth}
  \centering
  \includegraphics[width=0.4\textwidth, page=2]{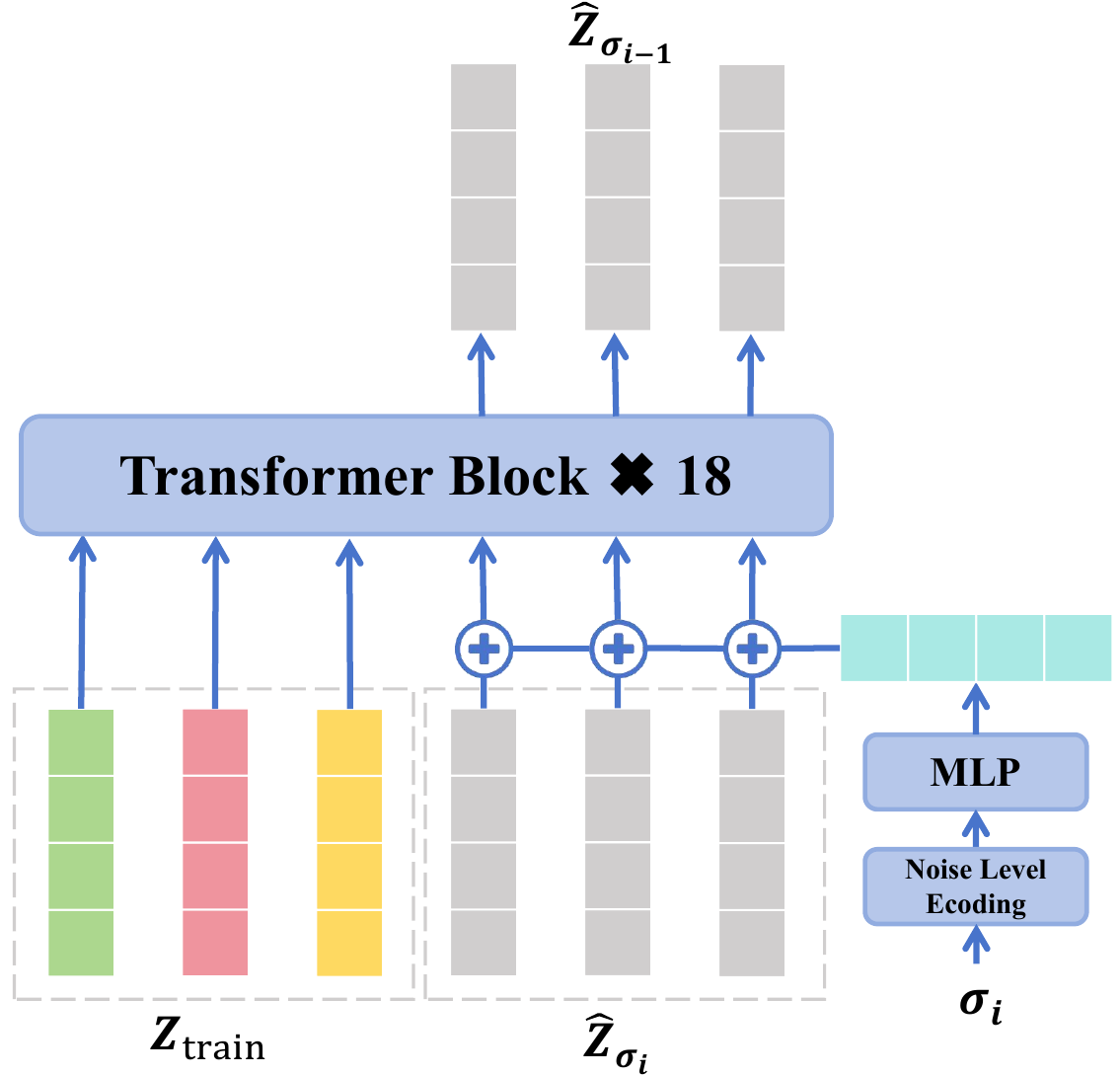}
  \caption{In-context learning model.}
  \label{fig:fig2}
\end{wrapfigure}

The in-context learning model $D_\theta$ in Fig \ref{fig:fig2} is an 18 layers Transformer and takes the concatenation of the clean embeddings $\mZ_\text{train}$ and noisy embeddings $\mZ_\sigma$ as input. 
Specifically, we employ a sample-wise attention mask: clean embeddings attend to each other but cannot attend to noisy embeddings; and each noisy embedding attends only to itself and clean embeddings, but not to other noisy embeddings. This allows the model to leverage the context for guided denoising while keeping the denoising processes of different noisy embeddings independent.
For the noise level $\sigma$, we apply sinusoidal encoding followed by a shallow MLP that projects it to 192 dimensions, and add it to the noisy query embeddings so that the model is aware of the current noise level.

\subsection{Dataset-Specific MLP Decoder}
After noise sampling and multi-step denoising, we obtain the generated latent embeddings $\hat{\mZ}$, which still needs to be mapped back to the concrete data space.
Because different datasets may exhibit distinct feature types, scales, and distributions, a shared decoder is insufficient. 
We train a lightweight MLP decoder separately for each dataset. 
This decoder is trained using the original samples $\mX_\text{train}$ and their corresponding latent embeddings 
$\mZ_\text{train}$. For any feature $f$, the corresponding MLP decoder maps the latent embedding $\mZ_\text{train}^{(f)}$ to either a scalar value or a probability distribution, and then computes the loss against the true $\mX_\text{train}^{(f)}$. Our optimization objective is as follows:
$$
\ell_\text{dec} = \sum_f\text{loss}_f\big(\text{MLPDecoder}_f(\mZ_\text{train}), \mX_\text{train}^{(f)}\big).
$$
Here, $\text{loss}_f$ denotes squared error for numerical features and the cross-entropy loss for categorical features.
For the latent embeddings $\hat{\mZ}$ generated by the diffusion model, the final generated data $\hat{\mX}$ can be obtained by feeding the corresponding embeddings into the associated decoder. Although this decoder is trained separately for each dataset at inference time, its lightweight architecture incurs only limited additional cost.

%% file: section_pretraining.tex
\section{Training Details}
\subsection{Training Data}
Following TabPFN, a general-purpose pretrained model for tabular data, we first train our model on large-scale synthetic data. Specifically, we use the open-source TabICL data generator to produce 5,000 batches of datasets. Each batch contains 512 datasets generated from different SCMs~\citep{pearl2009causality,poinsot2024learning}, where each dataset has between 128 and 2,048 samples and between 2 and 50 features. 
We reuse these 5,000 batches 10 times and randomly shuffle the columns, yielding 50,000 usable training batches in total.

For the second stage, we train on 72 real-world datasets from OpenML-CC18~\citep{oml-benchmarking-suites} and OpenML-CTR23~\citep{fischer2023openmlctr}, sampling rows and columns within the same ranges as the synthetic data. In each round, we sample one subset per dataset and group all 72 subsets into a single batch. Repeating this 1,000 times yields 1,000 batches, and reusing each batch 10 times gives 10,000 training batches in total.

\subsection{Training Process}

\begin{figure}[t]
  \centering
  \includegraphics[width=0.49\textwidth]{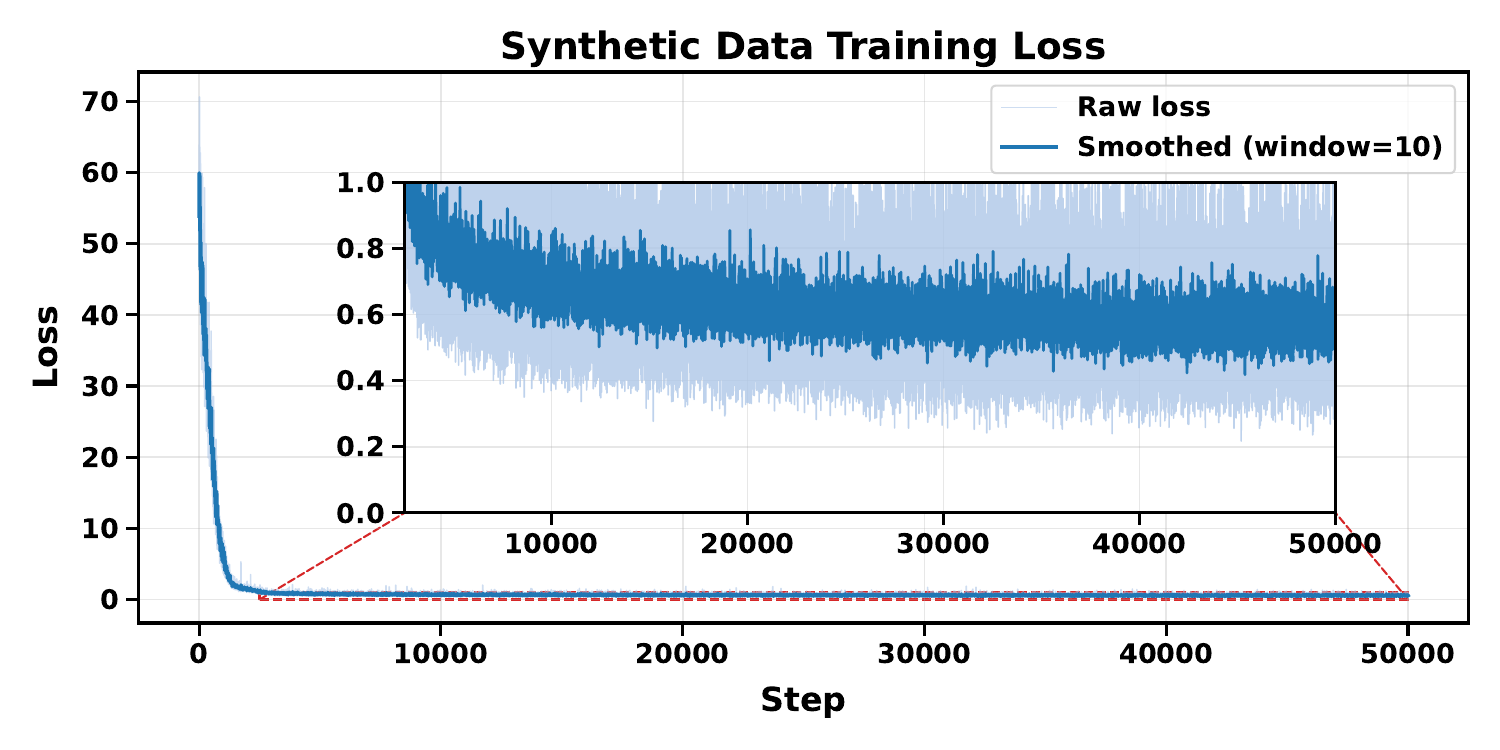}
  \hfill
  \includegraphics[width=0.49\textwidth]{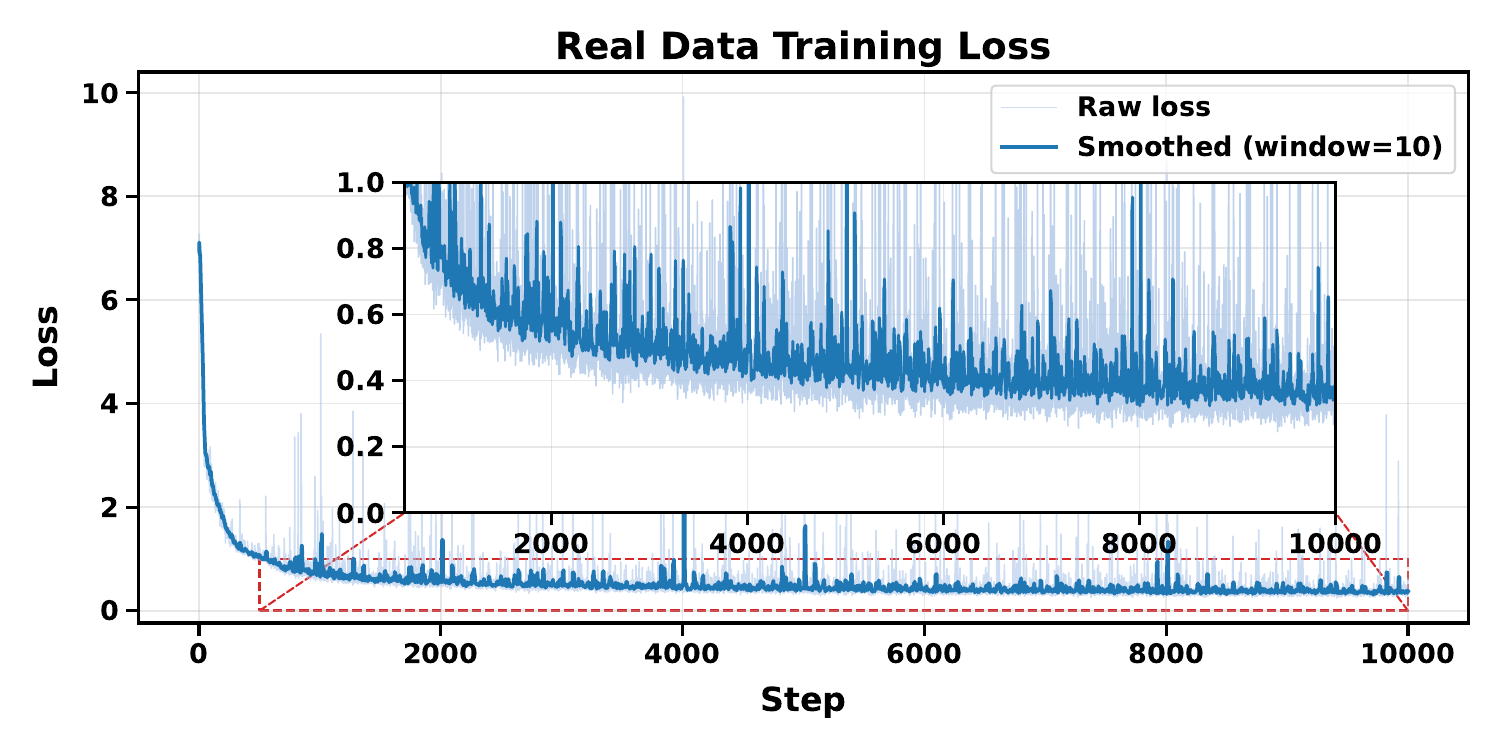}
  \caption{Loss curves during the two-stage training. The light blue curve shows the raw loss, and the dark blue curve shows the loss smoothed by a moving average with a window size of 10.}
  \label{fig:two}
\end{figure}
We first train the model on the aforementioned 50,000 synthetic training batches and then on 10,000 real-data training batches. 
For each dataset, we uniformly sample a context ratio between 0.5 and 0.8, and use the remaining portion for noising and reconstruction. The entire training process is conducted on a machine equipped with an Intel Xeon Platinum 8336C CPU and 4 NVIDIA RTX 4090 GPUs. It takes about 8 days in total, with the first 7 days using synthetic data and the last day using real data. The loss curve is shown in the figure ~\ref{fig:two}.
We use a cosine learning rate schedule with linear warmup. The initial learning rate is set to $2\times10^{-4}$ with a warmup proportion of 0.02, gradient clipping of 1.0.

%% file: section_experience.tex
\section{Experiment}
This section presents our experimental setup and results. We conduct experiments on 18 real-world datasets, consisting of 7 classification tasks and 11 regression tasks. The generated data are evaluated in terms of statistical fidelity, privacy preservation, and downstream model performance. We will introduce our experimental setup and results step by step.

\subsection{Datasets}
We use the tabular benchmark TALENT~\citep{ye2024closerlookdeeplearning,liu2024talenttabularanalyticslearning}, which contains 300 high-quality tabular datasets. We employ a subset provided by them, Tiny Talent, which consists of 45 datasets. After removing datasets with more than 50 features, we obtain 7 classification tasks and 11 regression tasks. Detailed information about the datasets is provided in Appendix~\ref{Datasets}.

To systematically evaluate various generative algorithms under different data budgets, we subsample from each dataset to construct training sets of sizes 16, 32, 64, 128, 256, and 512. 
These subsets are then used to train the generative models. 
For each training size, we repeat the experiment 5 times using 5 different random seeds. We save the generated data produced by each method under each training size for subsequent evaluation and comparison. This protocol allows us to assess how generation quality varies with the amount of available training data. 

For baselines that require standardized data, including {\sc TabGift}, we preprocess each dataset by applying Z-score normalization to continuous features and ordinal encoding to categorical features. This ensures that all methods operate on comparable input representations.

\subsection{Baseline Generators And Downstream Predictors}
We compare {\sc TabGift} with existing tabular data generation methods, including:
(i) the classical data augmentation method SMOTE~\citep{chawla2002smote};
(ii) the VAE-based method TVAE~\citep{xu2019modeling};
(iii) the GAN-based method CTGAN~\citep{xu2019modeling};
(iv) the diffusion-based method TabDDPM~\citep{kotelnikov2023tabddpm};
(v) the tree-based method ARF~\citep{watson2023adversarial};
(vi) a Bayesian network (BN)-based method~\citep{Ankan2024};
(vii) the normalizing flow method NFlow~\citep{durkan2019neural};
and (viii) TabEBM~\citep{margeloiu2024tabebm}.
For some of the methods that require hyperparameter tuning, we performed 20 optimization trials. Implementation details can be found in the Appendix~\ref{Baselines}. 

We select 6 representative downstream predictors:  XGBoost~\citep{chen2016xgboost}, Random Forest~\citep{breiman2001random},  TabPFN~\citep{hollmann2025accurate}, multi-layer perceptrons~\citep{rumelhart1986learning}, $k$-nearest neighbors~\citep{cover1967nearest}, and Linear Regression or Logistic Regression~\citep{cox1958regression}. For some downstream predictors, we perform hyperparameter tuning, while synthetic data is used to train the models.

\begin{table*}[t]
\centering
\caption{Relative performance (mean \(\pm\) std) of different generation methods under various training sizes. For classification, we report \(D_{\mathrm{syn}}/D_{\mathrm{real}}\) (gen/real, higher is better); for regression, we report \(D_{\mathrm{real}}/D_{\mathrm{syn}}\) (real/gen, higher is better). Standard deviations are shown as subscripts. The \textcolor{red}{best}, \textcolor{orange}{second-best}, and \textcolor{blue}{third-best} results among generative methods are highlighted. Real is shown as a reference and is not ranked.}
\label{tab:relative_performance}
\resizebox{\textwidth}{!}{%
\begin{tabular}{l|cccccc}
\toprule
\multicolumn{7}{c}{Classification (gen/real, higher is better)} \\
\cmidrule(lr){1-7}
Method & 16 & 32 & 64 & 128 & 256 & 512 \\
\cmidrule(lr){1-7}
\midrule
Real 
 & $1.0000_{\pm 0.0000}$ & $1.0000_{\pm 0.0000}$ & $1.0000_{\pm 0.0000}$ & $1.0000_{\pm 0.0000}$ & $1.0000_{\pm 0.0000}$ & $1.0000_{\pm 0.0000}$ \\
SMOTE 
 & $0.9927_{\pm 0.0329}$ & $0.9885_{\pm 0.0456}$ & \textcolor{blue}{$0.9990_{\pm 0.0249}$} & \textcolor{blue}{$0.9979_{\pm 0.0184}$} & \textcolor{red}{$0.9997_{\pm 0.0171}$} & \textcolor{red}{$0.9980_{\pm 0.0104}$} \\
TVAE 
 & $1.0035_{\pm 0.0933}$ & $0.9761_{\pm 0.1022}$ & $0.9502_{\pm 0.0926}$ & $0.9368_{\pm 0.0574}$ & $0.9310_{\pm 0.0772}$ & $0.8988_{\pm 0.0685}$ \\
CTGAN 
 & $0.9918_{\pm 0.1292}$ & $0.9611_{\pm 0.1227}$ & $0.9231_{\pm 0.1022}$ & $0.9019_{\pm 0.1083}$ & $0.9162_{\pm 0.0976}$ & $0.8842_{\pm 0.1055}$ \\
TabDDPM 
 & $0.9533_{\pm 0.0966}$ & $0.9046_{\pm 0.1158}$ & $0.9306_{\pm 0.1159}$ & $0.9176_{\pm 0.1201}$ & $0.9028_{\pm 0.1721}$ & $0.8925_{\pm 0.1348}$ \\
ARF 
 & $0.9720_{\pm 0.0648}$ & \textcolor{blue}{$0.9968_{\pm 0.0697}$} & $0.9702_{\pm 0.0510}$ & $0.9725_{\pm 0.0450}$ & $0.9627_{\pm 0.0273}$ & $0.9352_{\pm 0.0396}$ \\
BN 
 & \textcolor{blue}{$1.0128_{\pm 0.0710}$} & $0.9683_{\pm 0.0428}$ & \textcolor{orange}{$1.0007_{\pm 0.0747}$} & \textcolor{orange}{$0.9999_{\pm 0.0653}$} & \textcolor{blue}{$0.9907_{\pm 0.0663}$} & \textcolor{orange}{$0.9867_{\pm 0.0621}$} \\
NFlow 
 & $0.9754_{\pm 0.1095}$ & $0.9728_{\pm 0.1146}$ & $0.9411_{\pm 0.0907}$ & $0.9469_{\pm 0.0758}$ & $0.8954_{\pm 0.1029}$ & $0.8581_{\pm 0.0949}$ \\
TabEBM 
 & \textcolor{orange}{$1.0542_{\pm 0.0939}$} & \textcolor{orange}{$1.0006_{\pm 0.0663}$} & $0.9890_{\pm 0.0285}$ & $0.9961_{\pm 0.0279}$ & $0.9799_{\pm 0.0434}$ & \textcolor{blue}{$0.9612_{\pm 0.0500}$} \\
{\sc TabGift}  
 & \textcolor{red}{$1.0729_{\pm 0.1026}$} & \textcolor{red}{$1.0313_{\pm 0.0481}$} & \textcolor{red}{$1.0213_{\pm 0.0767}$} & \textcolor{red}{$1.0030_{\pm 0.0754}$} & \textcolor{orange}{$0.9923_{\pm 0.0693}$} & $0.9526_{\pm 0.0747}$ \\
\bottomrule
\end{tabular}%
}

\vspace{1.5em}

\resizebox{\textwidth}{!}{%
\begin{tabular}{l|cccccc}
\toprule
\multicolumn{7}{c}{Regression (real/gen, higher is better)} \\
\cmidrule(lr){1-7}
Method & 16 & 32 & 64 & 128 & 256 & 512 \\
\cmidrule(lr){1-7}
\midrule
Real 
 & $1.0000_{\pm 0.0000}$ & $1.0000_{\pm 0.0000}$ & $1.0000_{\pm 0.0000}$ & $1.0000_{\pm 0.0000}$ & $1.0000_{\pm 0.0000}$ & $1.0000_{\pm 0.0000}$ \\
SMOTE 
 & $0.9762_{\pm 0.1268}$ & $1.0451_{\pm 0.1128}$ & $1.0153_{\pm 0.2018}$ & $0.9618_{\pm 0.1627}$ & \textcolor{blue}{$0.9437_{\pm 0.1329}$} & \textcolor{blue}{$0.9192_{\pm 0.1604}$} \\
TVAE 
 & $1.0166_{\pm 0.1737}$ & $0.9802_{\pm 0.1310}$ & $0.9481_{\pm 0.1645}$ & $0.8414_{\pm 0.1437}$ & $0.8065_{\pm 0.1530}$ & $0.7953_{\pm 0.1801}$ \\
CTGAN 
 & $0.9205_{\pm 0.1799}$ & $0.9324_{\pm 0.1235}$ & $0.9575_{\pm 0.1921}$ & $0.8233_{\pm 0.1447}$ & $0.7936_{\pm 0.1422}$ & $0.7595_{\pm 0.1230}$ \\
TabDDPM 
 & $0.8655_{\pm 0.1907}$ & $0.8642_{\pm 0.2245}$ & $0.8477_{\pm 0.2062}$ & $0.8476_{\pm 0.2115}$ & $0.8133_{\pm 0.1857}$ & $0.8105_{\pm 0.1732}$ \\
ARF 
 & $1.0439_{\pm 0.1426}$ & $0.9598_{\pm 0.1484}$ & $0.9525_{\pm 0.1638}$ & $0.8960_{\pm 0.1582}$ & $0.8608_{\pm 0.1586}$ & $0.8295_{\pm 0.1541}$ \\
BN 
 & \textcolor{orange}{$1.1544_{\pm 0.1797}$} & \textcolor{blue}{$1.0476_{\pm 0.1590}$} & \textcolor{blue}{$1.0602_{\pm 0.1089}$} & \textcolor{blue}{$1.0058_{\pm 0.1260}$} & \textcolor{orange}{$1.0100_{\pm 0.0506}$} & \textcolor{orange}{$0.9722_{\pm 0.0646}$} \\
NFlow 
 & $0.9805_{\pm 0.1500}$ & $0.9167_{\pm 0.1437}$ & $0.8799_{\pm 0.1859}$ & $0.8039_{\pm 0.1358}$ & $0.7703_{\pm 0.1498}$ & $0.7580_{\pm 0.1476}$ \\
TabEBM 
 & \textcolor{blue}{$1.0836_{\pm 0.1518}$} & \textcolor{orange}{$1.1073_{\pm 0.1386}$} & \textcolor{orange}{$1.0870_{\pm 0.1195}$} & \textcolor{red}{$1.0297_{\pm 0.0892}$} & \textcolor{red}{$1.0303_{\pm 0.0691}$} & \textcolor{red}{$0.9772_{\pm 0.0751}$} \\
{\sc TabGift}  
 & \textcolor{red}{$1.1814_{\pm 0.1658}$} & \textcolor{red}{$1.1222_{\pm 0.1925}$} & \textcolor{red}{$1.1338_{\pm 0.2140}$} & \textcolor{orange}{$1.0247_{\pm 0.2000}$} & $0.9076_{\pm 0.1703}$ & $0.8359_{\pm 0.1992}$ \\
\bottomrule
\end{tabular}%
}
\end{table*}
\vspace{1.5em}
\subsection{Evaluation}
\subsubsection{Downstream Model Efficacy}
In the actual evaluation, downstream models are trained either using only the real training data or using synthetic data generated by different generation methods on top of the training data.
We denote the performance of models trained on real data as $D_{\mathrm{real}}$, and the performance of models trained on synthetic data as $D_{\mathrm{syn}}$.
We adopt a relative performance metric for evaluation. Specifically, the performance achieved using real data is normalized to 1.
For classification tasks, we use the Area Under the Receiver Operating Characteristic curve (AUROC) as the evaluation metric, where higher values indicate better performance. 
The relative performance is thus computed as $D_{\mathrm{syn}}$/$D_{\mathrm{real}}$.
For regression tasks, we instead use the Root Mean Squared Error (RMSE), where lower values indicate better performance. 
Accordingly, the relative performance is defined as $D_{\mathrm{real}}$/$D_{\mathrm{syn}}$. The detailed results are shown in Tab~\ref{tab:relative_performance}.

{\sc TabGift} consistently outperforms dataset-specific generative methods in the limited-context setting. Dataset-specific generative models are prone to memorizing training samples: with only a handful of examples, the empirical distribution is easily memorized, and the model struggles to distinguish generalizable structure from sample-specific patterns. 
In contrast, during ICL pretraining, the context is constantly changing and memorizing the context cannot reduce the pretraining loss. This removes the memorization incentive and forces the model to infer transferable distributional priors, enabling high-quality generation from limited context.
As the number of training samples increases, the hypothesis spaces of dataset-specific models expand with the data, and their empirical fitting ability gradually strengthens. 
In contrast, The performance {\sc TabGift} is mainly limited by the expressiveness of the pretrained prior and the information bottleneck of context compression, so its relative advantage narrows. Nevertheless, even at size 512, {\sc TabGift} remains in the upper half, indicating that its prior does not fail but that its marginal benefit decreases as the data volume grows.

\subsubsection{Statistical Fidelity And Privacy Protection}
\begin{figure}[t]
  \centering
  \includegraphics[width=0.9\textwidth]{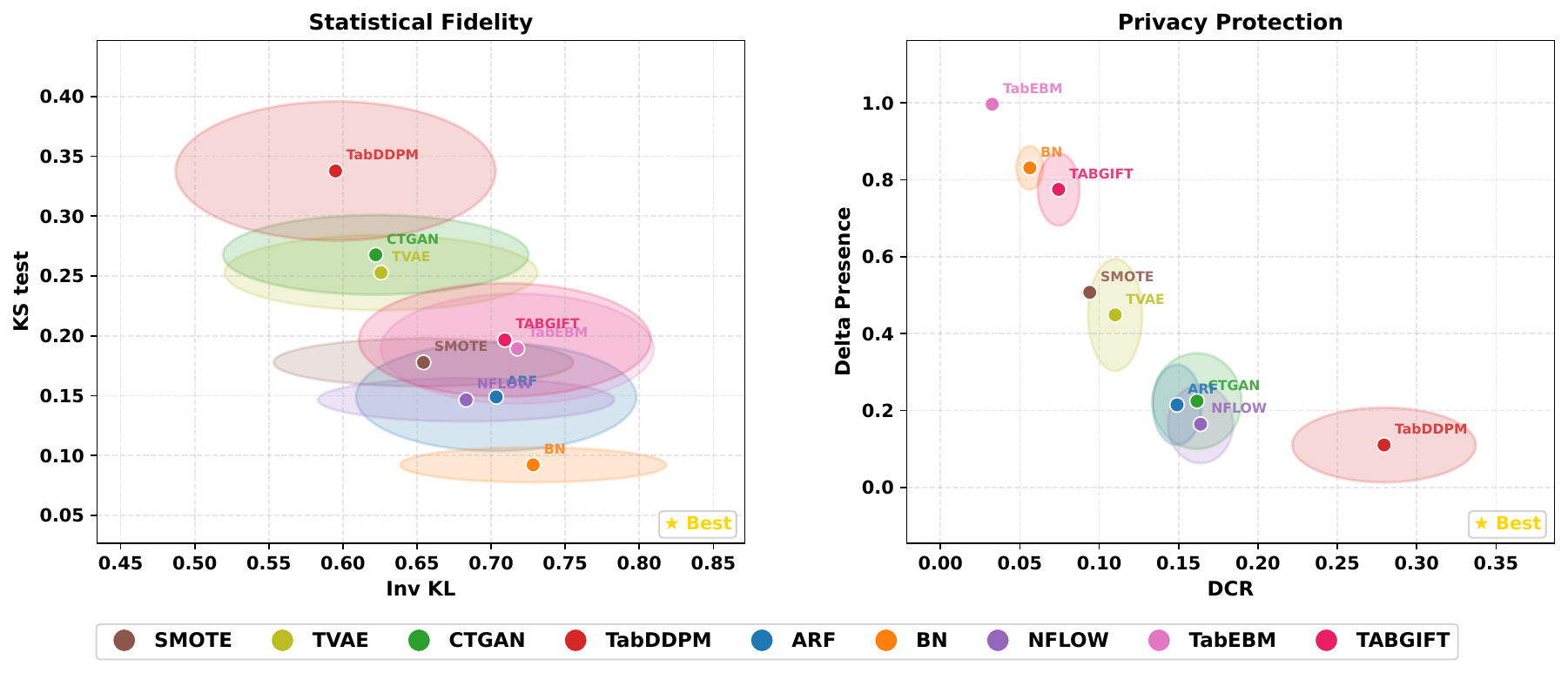}
  \vspace{-1em}
  \caption{Synthetic generator comparison on statistical fidelity (Inv KL, KS test) and privacy protection (DCR, Delta Presence). Shown for training size 16 across all datasets, mean over 5 seeds.}
  \label{fig:metric_s16}
\end{figure}

For statistical fidelity, we estimate the similarity between the training data and the synthetic data using: (i) Column-wise distribution matching: Kullback–Leibler divergence~\citep{csiszar1975divergence} is computed for categorical features, Wasserstein distance~\citep{monge1781memoire} for numerical features, and the results are inversely mapped to $[0,1]$ range and then averaged; (ii) Kolmogorov–Smirnov test~\citep{karson1968handbook} statistic; 
For privacy protection, we assess the risk of data leakage using the following two metrics (computed with respect to the sampled training set):
(i) Average Distance to Closest Record~\citep{zhao2021ctab}: larger values indicate greater differences between generated and training samples; (ii) $\delta$-presence~\citep{van2011encyclopedia}: smaller values indicate a lower probability of identifying training samples from the generated data. 
We report the results for training size 16 in Fig   ~\ref{fig:metric_s16}. {\sc TabGift} performs reasonably well on statistical fidelity but is slightly weaker on privacy preservation. Interestingly, we observe a tension between these two objectives: methods with strong statistical fidelity (e.g., BN, TabEBM, and {\sc TabGift}) tend to lag behind in privacy protection. A plausible explanation is that better distributional matching often requires preserving fine-grained statistics of the training data, which can bring generated samples closer to real records and thus increase privacy risk. This suggests that statistical fidelity and privacy preservation should be evaluated jointly rather than in isolation. 

\subsubsection{Running Time Analysis}
\begin{table*}[t]
\centering
\caption{Average runtime (seconds, mean) of different generation methods under various training sizes. For each size, we report training time, inference time, and total time.}
\label{tab:runtime}
\resizebox{\textwidth}{!}{%
\begin{tabular}{l|ccc|ccc|ccc|ccc|ccc|ccc}
\toprule
\multirow{2}{*}{Method} & \multicolumn{3}{c|}{16} & \multicolumn{3}{c|}{32} & \multicolumn{3}{c|}{64} & \multicolumn{3}{c|}{128} & \multicolumn{3}{c|}{256} & \multicolumn{3}{c}{512} \\
\cmidrule(lr){2-4} \cmidrule(lr){5-7} \cmidrule(lr){8-10} \cmidrule(lr){11-13} \cmidrule(lr){14-16} \cmidrule(lr){17-19}
 & Train & Infer & Total & Train & Infer & Total & Train & Infer & Total & Train & Infer & Total & Train & Infer & Total & Train & Infer & Total \\
\midrule
SMOTE 
 & 0.000 & 0.010 & 0.010 & 0.000 & 0.009 & 0.009 & 0.000 & 0.009 & 0.009 & 0.000 & 0.013 & 0.013 & 0.000 & 0.011 & 0.011 & 0.000 & 0.014 & 0.014 \\
TVAE 
 & 17.316 & 0.161 & 17.477 & 13.607 & 0.168 & 13.775 & 17.950 & 0.148 & 18.098 & 17.986 & 0.197 & 18.182 & 33.206 & 0.130 & 33.336 & 54.554 & 0.167 & 54.721 \\
CTGAN 
 & 30.704 & 0.128 & 30.832 & 32.919 & 0.129 & 33.049 & 50.764 & 0.113 & 50.878 & 55.871 & 0.101 & 55.972 & 69.029 & 0.095 & 69.123 & 94.568 & 0.097 & 94.665 \\
TabDDPM 
 & 29.110 & 333.780 & 362.890 & 29.493 & 135.185 & 164.677 & 28.566 & 71.906 & 100.472 & 26.288 & 67.191 & 93.479 & 24.295 & 20.538 & 44.834 & 26.733 & 19.629 & 46.362 \\
ARF 
 & 1.950 & 3.535 & 5.486 & 1.484 & 3.156 & 4.640 & 1.712 & 3.276 & 4.988 & 1.858 & 2.802 & 4.660 & 2.925 & 3.299 & 6.224 & 4.484 & 2.853 & 7.337 \\
BN 
 & 7.651 & 0.164 & 7.814 & 6.531 & 0.356 & 6.887 & 6.413 & 0.398 & 6.811 & 7.112 & 0.395 & 7.508 & 7.964 & 0.363 & 8.327 & 8.468 & 0.369 & 8.837 \\
NFlow 
 & 18.243 & 0.826 & 19.069 & 16.338 & 0.706 & 17.044 & 33.076 & 0.963 & 34.039 & 33.815 & 0.980 & 34.795 & 43.249 & 1.242 & 44.491 & 37.970 & 0.933 & 38.903 \\
TabEBM 
 & 0.000 & 27.150 & 27.150 & 0.000 & 27.009 & 27.009 & 0.000 & 27.923 & 27.923 & 0.000 & 28.793 & 28.793 & 0.000 & 29.490 & 29.490 & 0.000 & 28.098 & 28.098 \\
{\sc TabGift} 
 & 8.033 & 1.976 & 10.009 & 8.033 & 2.005 & 10.038 & 8.098 & 2.021 & 10.119 & 9.374 & 2.180 & 11.554 & 7.986 & 2.421 & 10.407 & 7.796 & 2.908 & 10.704 \\
\bottomrule
\end{tabular}%
}
\end{table*}

In tab~\ref{tab:runtime}, we generate 1,024 synthetic samples under different training sizes and compare the runtime of all methods on the Adult dataset. SMOTE is by far the fastest ($\approx$0.015s on average), since it only performs interpolation and requires no model training. 
TabEBM is also training-free, but it relies on multiple calls to TabPFN; its inference time alone ($\approx$28s) already exceeds the total time of {\sc TabGift} ($\approx$10.5s). 
Among methods that require training, {\sc TabGift} is only slower than ARF ($\approx$5.6s) and BN ($\approx$7.7s), while being substantially faster than TVAE ($\approx$25.9s), CTGAN ($\approx$55.8s), TabDDPM ($\approx$135.5s), and NFlow ($\approx$31.4s). The runtime of {\sc TabGift} is dominated by training the dataset-specific MLP decoder ($\approx$8.2s), and inference accounts for only about 20\% of the total time ($\approx$2.3s). This confirms that {\sc TabGift} achieves a favorable efficiency–quality trade-off, especially in the small-data regime.

%% file: appendix.tex
\section*{Appendix}
\section{Datasets}\label{Datasets}
\subsection{OpenML}
This set of datasets comes from the OpenML-CC18~\citep{oml-benchmarking-suites} classification benchmark and the OpenML-CTR23~\citep{fischer2023openmlctr} regression benchmark. OpenML-CC18 is a curated collection of 72 classification datasets, while OpenML-CTR23 is a curated collection of 36 regression datasets. Both are widely used in tabular learning research for standardized and reproducible evaluation. In our experiments, we select 72 datasets from these two benchmarks, covering both classification and regression tasks with diverse feature types, sample sizes, and dimensionalities. Detailed statistics of the selected datasets are provided in Tab~\ref{tab:datasets_reg} and Tab~\ref{tab:datasets_cls}.

\begin{table*}[!h]
\centering
\small
\caption{Statistics of the OpenML regression datasets.}
\label{tab:datasets_reg}
\resizebox{\textwidth}{!}{%
\begin{tabular}{lccc}
\toprule
Dataset & \#Features & \#Samples & \#Cat / \#Num \\
\midrule
abalone & 8 & 4,177 & 1 / 7 \\
airfoil\_self\_noise & 5 & 1,503 & 0 / 5 \\
auction\_verification & 7 & 2,043 & 6 / 1 \\
concrete\_compressive\_strength & 8 & 1,030 & 0 / 8 \\
physiochemical\_protein & 9 & 45,730 & 0 / 9 \\
superconductivity & 81 & 21,263 & 2 / 79 \\
geographical\_origin\_of\_music & 116 & 1,059 & 0 / 116 \\
naval\_propulsion\_plant & 14 & 11,934 & 1 / 13 \\
white\_wine & 11 & 4,898 & 0 / 11 \\
grid\_stability & 12 & 10,000 & 0 / 12 \\
video\_transcoding & 18 & 68,784 & 2 / 16 \\
wave\_energy & 48 & 72,000 & 0 / 48 \\
sarcos & 21 & 48,933 & 0 / 21 \\
california\_housing & 8 & 20,640 & 0 / 8 \\
cpu\_activity & 21 & 8,192 & 0 / 21 \\
diamonds & 9 & 53,940 & 3 / 6 \\
kin8nm & 8 & 8,192 & 0 / 8 \\
pumadyn32nh & 32 & 8,192 & 0 / 32 \\
socmob & 5 & 1,156 & 4 / 1 \\
kings\_county & 21 & 21,613 & 7 / 14 \\
brazilian\_houses & 9 & 10,692 & 6 / 3 \\
fps\_benchmark & 43 & 24,624 & 19 / 24 \\
health\_insurance & 11 & 22,272 & 9 / 2 \\
Moneyball & 14 & 1,232 & 4 / 10 \\
energy\_efficiency & 8 & 768 & 2 / 6 \\
student\_performance\_por & 30 & 649 & 29 / 1 \\
QSAR\_fish\_toxicity & 6 & 908 & 2 / 4 \\
cars & 17 & 804 & 16 / 1 \\
space\_ga & 6 & 3,107 & 0 / 6 \\
\bottomrule
\end{tabular}%
}
\end{table*}

\begin{table*}[!h]
\centering
\small
\caption{Statistics of the OpenML classification datasets.}
\label{tab:datasets_cls}
\resizebox{\textwidth}{!}{%
\begin{tabular}{lcccc@{\hspace{2em}}lcccc}
\toprule
Dataset & \#Classes & \#Features & \#Samples & \#Cat / \#Num
& Dataset & \#Classes & \#Features & \#Samples & \#Cat / \#Num \\
\midrule
kr-vs-kp & 2 & 36 & 3,196 & 36 / 0
& pc1 & 2 & 21 & 1,109 & 0 / 21 \\
letter & 26 & 16 & 20,000 & 16 / 0
& adult & 2 & 14 & 48,842 & 9 / 5 \\
balance-scale & 3 & 4 & 625 & 0 / 4
& Bioresponse & 2 & 1,776 & 3,751 & 834 / 942 \\
mfeat-factors & 10 & 216 & 2,000 & 22 / 194
& wdbc & 2 & 30 & 569 & 0 / 30 \\
mfeat-fourier & 10 & 76 & 2,000 & 0 / 76
& phoneme & 2 & 5 & 5,404 & 0 / 5 \\
breast-w & 2 & 9 & 699 & 0 / 9
& qsar-biodeg & 2 & 41 & 1,055 & 22 / 19 \\
mfeat-karhunen & 10 & 64 & 2,000 & 0 / 64
& wall-robot-navigation & 4 & 24 & 5,456 & 0 / 24 \\
mfeat-morphological & 10 & 6 & 2,000 & 0 / 6
& semeion & 10 & 256 & 1,593 & 256 / 0 \\
mfeat-zernike & 10 & 47 & 2,000 & 0 / 47
& ilpd & 2 & 10 & 583 & 1 / 9 \\
cmc & 3 & 9 & 1,473 & 8 / 1
& ozone-level-8hr & 2 & 72 & 2,534 & 0 / 72 \\
optdigits & 10 & 64 & 5,620 & 0 / 64
& cnae-9 & 9 & 856 & 1,080 & 856 / 0 \\
credit-approval & 2 & 15 & 690 & 9 / 6
& first-order-theorem-proving & 6 & 51 & 6,118 & 0 / 51 \\
credit-g & 2 & 20 & 1,000 & 17 / 3
& banknote-authentication & 2 & 4 & 1,372 & 0 / 4 \\
pendigits & 10 & 16 & 10,992 & 0 / 16
& blood-transfusion-service-center & 2 & 4 & 748 & 0 / 4 \\
diabetes & 2 & 8 & 768 & 0 / 8
& PhishingWebsites & 2 & 30 & 11,055 & 30 / 0 \\
spambase & 2 & 57 & 4,601 & 0 / 57
& cylinder-bands & 2 & 37 & 540 & 16 / 21 \\
splice & 3 & 60 & 3,190 & 60 / 0
& bank-marketing & 2 & 16 & 45,211 & 9 / 7 \\
vehicle & 4 & 18 & 846 & 1 / 17
& GesturePhaseSegmentationProcessed & 5 & 32 & 9,873 & 0 / 32 \\
electricity & 2 & 8 & 45,312 & 1 / 7
& har & 6 & 561 & 10,299 & 0 / 561 \\
satimage & 6 & 36 & 6,430 & 0 / 36
& dresses-sales & 2 & 12 & 500 & 11 / 1 \\
eucalyptus & 5 & 19 & 736 & 6 / 13
& texture & 11 & 40 & 5,500 & 0 / 40 \\
sick & 2 & 29 & 3,772 & 22 / 7
& connect-4 & 3 & 42 & 67,557 & 42 / 0 \\
vowel & 11 & 12 & 990 & 2 / 10
& MiceProtein & 8 & 77 & 1,080 & 0 / 77 \\
isolet & 26 & 617 & 7,797 & 0 / 617
& steel-plates-fault & 7 & 27 & 1,941 & 2 / 25 \\
analcatdata\_authorship & 4 & 70 & 841 & 29 / 41
& climate-model-simulation-crashes & 2 & 18 & 540 & 0 / 18 \\
analcatdata\_dmft & 5 & 4 & 797 & 4 / 0
& wilt & 2 & 5 & 4,839 & 0 / 5 \\
mnist\_784 & 10 & 784 & 70,000 & 150 / 634
& car & 4 & 6 & 1,728 & 6 / 0 \\
pc4 & 2 & 37 & 1,458 & 3 / 34
& segment & 7 & 16 & 2,310 & 0 / 16 \\
pc3 & 2 & 37 & 1,563 & 2 / 35
& Fashion-MNIST & 10 & 784 & 70,000 & 3 / 781 \\
jm1 & 2 & 21 & 10,885 & 0 / 21
& jungle\_chess\_2pcs\_raw\_endgame\_complete & 3 & 6 & 44,819 & 6 / 0 \\
kc2 & 2 & 21 & 522 & 1 / 20
& numerai28.6 & 2 & 21 & 96,320 & 0 / 21 \\
kc1 & 2 & 21 & 2,109 & 1 / 20
& Devnagari-Script & 46 & 1,024 & 92,000 & 240 / 784 \\
 & & & &
& CIFAR\_10 & 10 & 3,072 & 60,000 & 0 / 3,072 \\
 & & & &
& Internet-Advertisements & 2 & 1,558 & 3,279 & 1,555 / 3 \\
 & & & &
& dna & 3 & 180 & 3,186 & 180 / 0 \\
 & & & &
& churn & 2 & 20 & 5,000 & 4 / 16 \\
\bottomrule
\end{tabular}%
}
\end{table*}

\subsection{Talent}
TALENT~\citep{liu2024talenttabularanalyticslearning} is a comprehensive benchmark and toolbox for tabular data learning, providing a large collection of high-quality tabular datasets together with standardized pipelines for data preprocessing, model training, and evaluation. It includes 300 tabular datasets spanning classification and regression tasks, and provides a smaller curated subset, TALENT-tiny, for efficient and reproducible evaluation. In our experiments, we select 18 datasets from TALENT-tiny as our test set, covering both classification and regression tasks. Detailed statistics of the selected datasets are provided in Tab~\ref{tab:datasets_cls} and Tab~\ref{tab:datasets_reg}.

\begin{table*}[!h]
\centering
\small
\caption{Statistics of the TALENT classification datasets.}
\label{tab:talent_cls}
\resizebox{\textwidth}{!}{%
\begin{tabular}{lccccc}
\toprule
Dataset & \#Rows & \#Features & \#Cat & \#Num & \#Classes \\
\midrule
FOREX\_audchf-day-High & 1,833 & 10 & 0 & 10 & 2 \\
FOREX\_audsgd-hour-High & 43,825 & 10 & 0 & 10 & 2 \\
FOREX\_cadjpy-hour-High & 43,825 & 10 & 0 & 10 & 2 \\
Gender\_Gap\_in\_Spanish\_WP & 4,746 & 13 & 0 & 13 & 3 \\
adult & 45,222 & 14 & 8 & 6 & 2 \\
moons & 512 & 2 & 0 & 2 & 2 \\
pc3 & 1,563 & 37 & 1 & 36 & 2 \\
taiwanese\_bankruptcy\_prediction & 6,819 & 95 & 2 & 93 & 2 \\
website\_phishing & 1,353 & 9 & 9 & 0 & 3 \\
\bottomrule
\end{tabular}%
}
\end{table*}

\begin{table*}[!h]
\centering
\small
\caption{Statistics of the TALENT regression datasets.}
\label{tab:talent_reg}
\resizebox{\textwidth}{!}{%
\begin{tabular}{lccccc}
\toprule
Dataset & \#Rows & \#Features & \#Cat & \#Num & \#Unique Targets \\
\midrule
Ailerons & 13,750 & 40 & 12 & 28 & 35 \\
CookbookReviews & 18,182 & 7 & 1 & 6 & 6 \\
IEEE80211aa-GATS & 4,046 & 27 & 2 & 25 & 3,407 \\
Large-scale\_Wave\_Energy\_Farm\_Sydney\_49 & 17,964 & 99 & 0 & 99 & 4,816 \\
Superconductivty & 21,197 & 81 & 2 & 79 & 3,007 \\
VulNoneVul & 5,692 & 16 & 0 & 16 & 2 \\
archive2 & 1,143 & 12 & 1 & 11 & 1,143 \\
bank8FM & 8,192 & 8 & 1 & 7 & 7,913 \\
communities\_and\_crime & 1,994 & 102 & 2 & 100 & 98 \\
compass\_reg & 16,644 & 17 & 8 & 9 & 2 \\
fried & 40,768 & 10 & 0 & 10 & 17,396 \\
house\_16H\_reg & 22,784 & 16 & 0 & 16 & 2,045 \\
kin8nm & 8,192 & 8 & 0 & 8 & 8,191 \\
\bottomrule
\end{tabular}%
}
\end{table*}

\section{Baselines}\label{Baselines}
Below is a detailed introduction and implementation of the baseline methods used in this paper:
\begin{itemize}
    \item \textbf{SMOTE}~\citep{chawla2002smote}
    
    \quad\textbf{Introduction}: Synthetic Minority Over-sampling Technique generates new samples for minority classes via interpolation. It helps balance imbalanced datasets and improves classifier performance on underrepresented classes.
    
    \quad\textbf{Implementation}: Use the open-source implementation of SMOTE from the python library Imbalanced-learn~\citep{lemaavztre2017imbalanced}. \url{https://imbalanced-learn.org/stable/}.

    \quad\textbf{Hyperparameter Space}:

    \quad\quad\textbf{k}: integer, choices $\{1, 3, 5\}$
    
    \item \textbf{TVAE}~\citep{xu2019modeling}: 
    
    \quad\textbf{Introduction}: A Variational Autoencoder designed for tabular data that captures feature dependencies in a latent space. It generates realistic synthetic samples while preserving correlations among columns.

    \quad\textbf{Implementation}: Use the open-source implementation in the python library Synthcity~\citep{qian2023synthcity}. \url{https://github.com/vanderschaarlab/synthcity/tree/main}.

    \quad\textbf{Hyperparameter Space}:

    \quad\quad \textbf{n\_iter}: integer, choices $\{100, 200, 300, 400, 500\}$
    
    \quad\quad \textbf{lr}: float, choices $\{0.0001, 0.0002, 0.001\}$
    
    \quad\quad \textbf{decoder\_n\_layers\_hidden}: integer, range $[1, 5]$
    
    \quad\quad \textbf{weight\_decay}: float, choices $\{0.0001, 0.001\}$
    
    \quad\quad \textbf{batch\_size}: integer, choices $\{64, 128, 256, 512\}$
    
    \quad\quad \textbf{n\_units\_embedding}: integer, range $[50, 500]$ step 50
    
    \quad\quad \textbf{decoder\_n\_units\_hidden}: integer, range $[50, 500]$ step 50
    
    \quad\quad \textbf{decoder\_nonlin}: categorical, choices $\{\text{elu}, \text{leaky\_relu}, \text{relu}, \text{tanh}\}$
    
    \quad\quad \textbf{decoder\_dropout}: float, range $[0.0, 0.2]$
    
    \quad\quad \textbf{encoder\_n\_layers\_hidden}: integer, range $[1, 5]$
    
    \quad\quad \textbf{encoder\_n\_units\_hidden}: integer, range $[50, 500]$ step 50
    
    \quad\quad \textbf{encoder\_nonlin}: categorical, choices $\{\text{elu}, \text{leaky\_relu}, \text{relu}, \text{tanh}\}$
    
    \quad\quad \textbf{encoder\_dropout}: float, range $[0.0, 0.2]$

    \item \textbf{CTGAN}~\citep{xu2019modeling}: 
    
    \quad\textbf{Introduction}: A GAN-based model for tabular data that uses conditional sampling to handle discrete columns. It effectively models mixed-type tables and generates high-quality synthetic data.

    \quad\textbf{Implementation}: Use the open-source implementation in the python library Synthcity.

    \quad\textbf{Hyperparameter Space}:

    \quad\quad \textbf{generator\_n\_layers\_hidden}: integer, range $[1, 4]$
    
    \quad\quad \textbf{generator\_n\_units\_hidden}: integer, range $[50, 150]$, step 50
    
    \quad\quad \textbf{generator\_nonlin}: categorical, choices $\{\text{elu}, \text{leaky\_relu}, \text{relu}, \text{tanh}\}$
    
    \quad\quad \textbf{n\_iter}: integer, range $[100, 1000]$, step 100
    
    \quad\quad \textbf{generator\_dropout}: float, range $[0.0, 0.2]$
    
    \quad\quad \textbf{discriminator\_n\_layers\_hidden}: integer, range $[1, 4]$
    
    \quad\quad \textbf{discriminator\_n\_units\_hidden}: integer, range $[50, 150]$, step 50
    
    \quad\quad \textbf{discriminator\_nonlin}: categorical, choices $\{\text{elu}, \text{leaky\_relu}, \text{relu}, \text{tanh}\}$
    
    \quad\quad \textbf{discriminator\_n\_iter}: integer, range $[1, 5]$
    
    \quad\quad \textbf{discriminator\_dropout}: float, range $[0.0, 0.2]$
    
    \quad\quad \textbf{lr}: float, choices $\{0.0001, 0.0002, 0.001\}$
    
    \quad\quad \textbf{weight\_decay}: float, choices $\{0.0001, 0.001\}$
    
    \quad\quad \textbf{batch\_size}: integer, choices $\{100, 200, 500\}$
    
    \quad\quad \textbf{encoder\_max\_clusters}: integer, range $[2, 20]$
    
    \item \textbf{TabDDPM}~\citep{kotelnikov2023tabddpm}: 

    \quad\textbf{Introduction}: A diffusion-based generative model that denoises samples in data or latent space. It gradually transforms noise into realistic tabular samples using a learned denoising process.
    
    \quad\textbf{Implementation}: Use the open-source implementation in the python library Synthcity.

    \quad\textbf{Hyperparameter Space}:

    \quad\quad \textbf{lr}: log-uniform, range $[10^{-5}, 0.1]$
    
    \quad\quad \textbf{batch\_size}: integer log-uniform, range $[256, 4096]$
    
    \quad\quad \textbf{num\_timesteps}: integer, range $[10, 1000]$
    
    \quad\quad \textbf{n\_iter}: integer log-uniform, range $[1000, 10000]$
    
    \item \textbf{ARF}~\citep{watson2023adversarial}: 

    \quad\textbf{Implementation}: Use the open-source implementation in the python library Synthcity.
    
    \quad\textbf{Introduction}: Adversarial Random Forests is a non-parametric method for density estimation and synthetic data generation. It can model intricate feature interactions without assuming a parametric form.

    \quad\textbf{Hyperparameter Space}:

    \quad\quad \textbf{num\_trees}: integer, range $[10, 100]$, step 10
    
    \quad\quad \textbf{delta}: float, range $[0.0, 0.5]$
    
    \quad\quad \textbf{max\_iters}: integer, range $[1, 5]$
    
    \quad\quad \textbf{early\_stop}: categorical, choices $\{\text{False}, \text{True}\}$
    
    \quad\quad \textbf{min\_node\_size}: integer, range $[2, 20]$, step 2

    \item \textbf{BN}~\citep{Ankan2024}: 
    
    \quad\textbf{Introduction}: Bayesian Network-based generator models variable dependencies via a directed acyclic graph. New samples are generated by sampling from the joint probability distribution encoded by the network.

    \quad\textbf{Implementation}: Use the open-source implementation in the python library Synthcity.

    \quad\textbf{Hyperparameter Space}:

    \quad\quad \textbf{struct\_learning\_search\_method}: categorical, choices $\{\text{hillclimb}, \text{pc}, \text{tree\_search}\}$
    
    \quad\quad \textbf{struct\_learning\_score}: categorical, choices $\{\text{bdeu}, \text{bds}, \text{bic}, \text{k2}\}$

    \item \textbf{NFlow}~\citep{durkan2019neural}: 

    \quad\textbf{Introduction}: Neural Flow-based generator leverages normalizing flows to model complex joint distributions. It transforms a simple base distribution into the target data distribution through a sequence of invertible transformations, enabling efficient sampling and exact likelihood estimation.

    \quad\textbf{Implementation}: Use the open-source implementation in the Python library Synthcity.

    \quad\textbf{Hyperparameter Space}:

    \quad\quad \textbf{n\_iter}: integer, range $[100, 5000]$, step 100

    \quad\quad \textbf{n\_layers\_hidden}: integer, range $[1, 10]$

    \quad\quad \textbf{n\_units\_hidden}: integer, range $[10, 100]$

    \quad\quad \textbf{batch\_size}: categorical, choices $\{32, 64, 128, 256, 512\}$

    \quad\quad \textbf{dropout}: float, range $[0.0, 0.2]$

    \quad\quad \textbf{batch\_norm}: categorical, choices $\{\text{False}, \text{True}\}$

    \quad\quad \textbf{lr}: categorical, choices $\{0.0001, 0.0002, 0.001\}$

    \quad\quad \textbf{linear\_transform\_type}: categorical, choices $\{\text{lu}, \text{permutation}, \text{svd}\}$

    \quad\quad \textbf{base\_transform\_type}: categorical, choices $\{\text{affine-autoregressive}, \text{affine-coupling},$
    
    \quad\quad $\text{quadratic-autoregressive}, \text{quadratic-coupling}, \text{rq-autoregressive}, \text{rq-coupling}\}$
    
    \item \textbf{TabEBM}~\citep{margeloiu2024tabebm}: 

    \quad\textbf{Implementation}: Use the open-source implementation in the python library TabEBM. \url{https://github.com/andreimargeloiu/TabEBM}.
    
    \quad\textbf{Introduction}: Combines TabPFN with energy-based models to synthesize high-quality tabular data. This approach balances flexibility and accuracy by leveraging energy-based likelihood estimation.
\end{itemize}

\section{Downstream model}\label{Downstream}
Below is a detailed introduction and implementation of the downstream model used in this paper:
\begin{itemize}
    \item \textbf{XGBoost}~\citep{chen2016xgboost}
    
    \quad\textbf{Introduction}: A scalable and efficient gradient boosting framework for supervised learning. It constructs an ensemble of decision trees to optimize prediction performance while controlling overfitting.  

    \quad\textbf{Implementation}: Use the open-source implementation in the python library xgboost. \url{https://xgboost.ai/}

    \quad\textbf{Hyperparameter Space}:

    \quad\quad \textbf{alpha}: float, $[10^{-8}, 100]$ (log scale)
    
    \quad\quad \textbf{colsample\_bylevel}: float, $[0.5, 1.0]$
    
    \quad\quad \textbf{colsample\_bytree}: float, $[0.5, 1.0]$
    
    \quad\quad \textbf{gamma}: float, $[10^{-8}, 100]$ (log scale)
    
    \quad\quad \textbf{lambda}: float, $[10^{-8}, 100]$ (log scale)
    
    \quad\quad \textbf{learning\_rate}: float, $[10^{-4}, 0.3]$ (log scale)
    
    \quad\quad \textbf{max\_depth}: integer, $[3, 10]$
    
    \quad\quad \textbf{min\_child\_weight}: float, $[10^{-2}, 100]$ (log scale)
    
    \quad\quad \textbf{subsample}: float, $[0.5, 1.0]$
    
    \item \textbf{Random Forest}~\citep{breiman2001random}
    
    \quad\textbf{Introduction}: An ensemble method that builds multiple decision trees and aggregates their predictions. It is robust to overfitting and can capture complex feature interactions.  

    \quad\textbf{Implementation}: Use the open-source implementation in the python library scikit-learn~\citep{pedregosa2011scikit}. \url{https://scikit-learn.org/stable/index.html}

    \quad\textbf{Hyperparameter Space}:

    \quad\quad \textbf{min\_samples\_split}: integer, $[2, 10]$
    \quad\quad \textbf{min\_samples\_leaf}: integer, $[1, 10]$
    
    \item \textbf{TabPFN}~\citep{hollmann2025accurate}
    
    \quad\textbf{Introduction}: A tabular foundation model that performs in-context learning for classification and regression tasks. It can estimate predictive distributions without explicit training on the downstream dataset.  

    \quad\textbf{Implementation}: Use the open-source implementation in the python library tabpfn. \url{https://github.com/PriorLabs/TabPFN}.
    
    \item \textbf{Linear Regression or Logistic Regression}~\citep{cox1958regression}
    
    \quad\textbf{Introduction}: Classical parametric models for regression and classification. Linear regression models continuous outcomes, while logistic regression models binary outcomes using a sigmoid function. 

    \quad\textbf{Implementation}: Use the open-source implementation in the python library scikit-learn.

    \item \textbf{K-Nearest Neighbors (KNN)}~\citep{cover1967nearest}

    \quad\textbf{Introduction}: A non-parametric method for classification and regression that predicts the output of a sample based on the majority label (for classification) or average value (for regression) of its $k$ nearest neighbors in the feature space.

    \quad\textbf{Implementation}: Use the open-source implementation in the python library scikit-learn. \url{https://scikit-learn.org/stable/modules/generated/sklearn.neighbors.KNeighborsClassifier.html}

    \quad\textbf{Hyperparameters}:

    \quad\quad \textbf{n\_neighbors}: integer, range $[1, 20]$
    
    \quad\quad \textbf{weights}: categorical, choices $\{\text{uniform}, \text{distance}\}$
    
    \quad\quad \textbf{p}:  integer, range $[1, 3]$

    \item \textbf{Multi-Layer Perceptron (MLP)}~\citep{rumelhart1986learning}

    \quad\textbf{Introduction}: A feedforward neural network with one or more hidden layers that can model complex non-linear relationships for regression or classification tasks. Uses backpropagation to learn network weights.

    \quad\textbf{Implementation}: Use the open-source implementation in the python library scikit-learn.

    \quad\textbf{Hyperparameters}:
    
    \quad\quad \textbf{hidden\_dim}: integer, range $[64, 512]$
    
    \quad\quad \textbf{num\_layers}: integer, range $[1, 3]$
    
    \quad\quad \textbf{dropout}: float, range $[0.0, 0.5]$
    
    \quad\quad \textbf{lr}: float, range $[1e-4, 1e-2]$ (log scale)
    
    \quad\quad \textbf{weight\_decay}: float, range $[1e-6, 1e-2]$ (log scale)
    
    \quad\quad \textbf{batch\_size}: integer, choices $\{32, 64, 128\}$
    
    \quad\quad \textbf{num\_epochs}: integer, range $[50, 500]$

\end{itemize}